\documentclass[10pt,twocolumn,letterpaper]{article}

\usepackage{iccv}
\usepackage{times}
\usepackage{epsfig}
\usepackage{graphicx}
\usepackage{amsmath}
\usepackage{amssymb}
\usepackage{makecell}
\usepackage{balance}
  \newcommand\figcaption{\def\@captype{figure}\caption}
  \newcommand\tabcaption{\def\@captype{table}\caption}
\makeatother
\usepackage[pagebackref=true,breaklinks=true,letterpaper=true,colorlinks,bookmarks=false]{hyperref}

\iccvfinalcopy % *** Uncomment this line for the final submission

\def\iccvPaperID{7556} % *** Enter the ICCV Paper ID here

\ificcvfinal\fi

\begin{document}

%%%%%%%%% TITLE
\title{GCC: Generative Calibration Clustering}

\author{Haifeng Xia$^{1}$, Hai Huang$^{2}$, Zhengming Ding$^{1}$ \\
$^{1}$Department of Computer Science, Tulane University, $^{2}$Google, Inc\\
{\tt\small \{hfxia,zding1\}@tulane.edu, haih@google.com}
}

% \author{Haifeng Xia\\
% Institution1\\
% Institution1 address\\
% {\tt\small firstauthor@i1.org}
% % For a paper whose authors are all at the same institution,
% % omit the following lines up until the closing ``}''.
% % Additional authors and addresses can be added with ``\and'',
% % just like the second author.
% % To save space, use either the email address or home page, not both
% \and
% Second Author\\
% Institution2\\
% First line of institution2 address\\
% {\tt\small secondauthor@i2.org}
% }

\maketitle
% Remove page # from the first page of camera-ready.
\ificcvfinal\thispagestyle{empty}\fi

%%%%%%%%% ABSTRACT
\begin{abstract}
   Deep clustering as an important branch of unsupervised representation learning focuses on embedding semantically similar samples into the identical feature space. This core demand inspires the exploration of contrastive learning and subspace clustering. However, these solutions always rely on the basic assumption that there are sufficient and category-balanced samples for generating valid high-level representation. This hypothesis actually is too strict to be satisfied for real-world applications. To overcome such a challenge, the natural strategy is utilizing generative models to augment considerable instances. How to use these novel samples to effectively fulfill clustering performance improvement is still difficult and under-explored. In this paper, we propose a novel Generative Calibration Clustering (GCC) method to delicately incorporate feature learning and augmentation into clustering procedure. First, we develop a discriminative feature alignment mechanism to discover intrinsic relationship across real and generated samples. Second, we design a self-supervised metric learning to generate more reliable cluster assignment to boost the conditional diffusion generation. Extensive experimental results on three benchmarks validate the effectiveness and advantage of our proposed method over the state-of-the-art methods.
\end{abstract}

%%%%%%%%% BODY TEXT
\section{Introduction}

As a fundamental research topic, image clustering attracts more attentions in machine learning and computer vision community \cite{caron2020unsupervised}. Since it brings abundant benefits to the downstream tasks, clustering technique has been successfully applied into many practical scenarios such as object recognition \cite{shen2021structure}, and scene understanding \cite{ng2006medical}. Its primary expectation is to group semantically similar images into the identical cluster without any external supervision. However, due to that images are lying in the sparse high-dimensional space, it becomes difficult to reach such a vision via the similarity estimation over the original input.

The earlier explorations \cite{macqueen1965some,zelnik2004self} typically adopt shallow models to transform images into low-dimensional feature space, where different similarity metrics are developed to measure sample-wise relationship. Although they achieve appealing performance in several simple benchmarks such as MNIST, the low-level representations likely involve considerable task-irrelevant semantics, reducing their discriminative property, especially for natural images. 

Benefited from deep neural network extracting sufficient semantics\cite{xia2020embedded,jing2023marginalized}, this issue is significantly alleviated by deep clustering solutions. The initial attempts \cite{yang2016joint,caron2018deep} iteratively optimize representation learning and clustering to produce mutual benefits for each other. However, this learning strategy needs conduct offline clustering over the entire dataset. It easily suffers from bottleneck when processing large-scale dataset or streaming data in the realistic applications. The obstruction motivates \cite{peng2019deep} to introduce the additional clustering head over the high-level representations. In this case, this component independently assigns each data point into the corresponding cluster. Later on, contrastive learning \cite{chen2020simple} achieves promising performance in unsupervised representation learning. The recent deep clustering algorithms \cite{dang2021doubly} almost adopt it to initialize extensive network parameters and further fine-tune the model by advancing the contrastive loss function. Unfortunately, when collecting negative pairs, it is likely to incorrectly divide positive ones into this type. Specifically, the half of negative pairs are supposed to be positive ones when there only exist two categories. And this issue also occurs in imbalanced category scenario. This situation results in the expanded intra-class distance and less discriminative feature.

The intuitive solution is precisely estimating sample-wise relationship and avoiding the negative effect of misjudgment. But it becomes very tough without label information. To eliminate this limitation, this paper explores conditional diffusion model to augment the desired images with pseudo annotations. With them, the clustering issue seems to be converted into a supervised learning task. Actually, the conversion exists obvious risk since the generation model may fail to perfectly approximate the realistic data distribution. Then the challenge is how to effectively associate generative images with the real ones and delivery novel supervised signals to bootstrap clustering. Inspired by this observation, we propose a novel method named Generative Calibration Clustering (GCC) with clustering and generation branches. For the former, our GCC utilizes the annotation information of generative images to build category-wise matching adjusting the distribution of feature space, and develops a novel reliable metric to emphasis discriminative semantics of novel samples. The latter receives pseudo labels of clustering head to gradually learn intrinsic class pattern and refine generative images. Moreover, to associate two image sets, our GCC considers the class centers of generative images as the anchors to calibrate that of realistic images. The main contributions are summarized as three folds:
\begin{itemize}
    \item This work explores conditional diffusion generation model to fulfill better clustering. For it, the important step is to build their association via precise calibration of class centers across realistic and augmented images.
    \item To obtain discriminative representations, our GCC estimates sample-wise intrinsic relation and develops a metric to perform reliable self-supervised learning over the generative images.
    \item Empirical analysis and experimental results on three benchmarks not only clearly explain the working mechanism of our GCC but also verify its effectiveness on solving clustering task.
\end{itemize}

\section{Related Works}
\textbf{Deep Clustering.} Image Clustering aims to group samples with the same object into the identical cluster. Achieving this vision needs to learn discriminative and robust features from high-dimensional visual signals. Fortunately, deep neural network utilizes hierarchical structure to capture valuable complicated semantics, which promotes the exploration of deep clustering \cite{asano2019self,peng2022xai,xia2020hgnet}. The pioneers \cite{yang2016joint,caron2018deep} build representation learning and clustering branches and iteratively optimize them. Later, due to the introduce of clustering head, the end-to-end training replaces the iterative one and obtains specific features with the concrete demand \cite{li2021contrastive}. Recently, inspired by the success of contrastive loss in unsupervised representation learning\cite{xia2021adaptive}, most deep clustering methods \cite{park2021improving,dang2021doubly} adopt it to initialize network parameters and control intra-class and inter-class distance. Different from them, our proposed GCC utilizes generation model to instruct feature learning for better clustering.

\textbf{Diffusion Generation Model.} As an image synthesis method, diffusion probabilistic models \cite{sohl2015deep} have achieved promising performance in density estimation and quality of image generation. The achievements result from its powerful fitting ability to the added noise by using U-Net architecture \cite{kingma2021variational}. This ability helps model accurately recover real data distribution from the random Gaussian noise. To further liberate its generation potentiality, \cite{ho2020denoising} introduces weight adjustment into objective function and \cite{rombach2022high} compresses image into the latent feature space where diffusion and denoising procedures are conducted. In addition, the advanced sampling mechanism \cite{kong2021fast,san2021noise} and hierarchical solution \cite{vahdat2021score,ho2022cascaded} are presented to solve expensive computational resource. Compared with these works, we focus on how to generate our desired images with the guidance of annotation under the diffusion model framework and how to utilize these augmented samples to obtain better clustering.

\section{Proposed Method}

\begin{figure*}[t]
    \centering
    \includegraphics[width=1\textwidth]{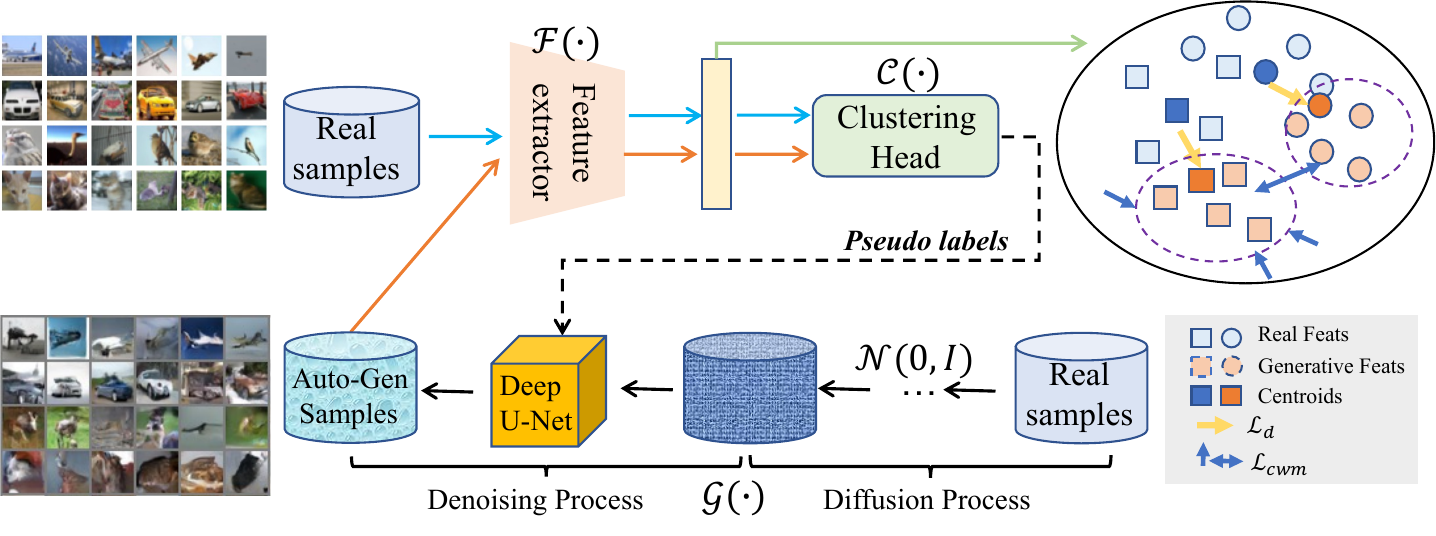}\vspace{-3mm}
    \caption{Overview of our Generative Calibration Clustering (GCC) framework. Our GCC includes clustering and image generation branches and utilizes their collaboration to learning discriminative representations. Concretely, $\mathcal{L}_{d}$ associates features of real and generative samples, while $\mathcal{L}_{cwm}$ and $\mathcal{L}_{ml}$ explore the labels of novel images to adjust feature distribution and emphasize important semantics.}\vspace{-3mm}
    \label{framework}
\end{figure*}

\subsection{Framework Overview} 

Clustering performance highly relies on the feature discrimination and robustness. To reach this, we propose a novel Generative Calibration Clustering (GCC) method (Figure \ref{framework}) to delicately incorporate them into clustering procedure, which adopts the intrinsic relationship across various samples and explores a self-supervised metric to conduct reliable cluster assignment. 

Specifically, we follow the mainstream deep clustering methods \cite{dang2021doubly,ma2022locally} with stacked network architectures to transform high-dimensional images $\mathbf{X}_{i}\in \mathbb{R}^{c\times h\times w}$ into the compact feature space via $\mathcal{F}(\mathbf{X}_{i})\in \mathbb{R}^{d}$ where $c$, $h$ and $w$ are separately channel number, height and width of image, and $d$ is the dimension of high-level feature. To enhance the feature robustness and discrimination of $\mathcal{F}(\cdot)$, we are the first to explore the diffusion model $\mathcal{G}(\cdot)$ to generate fake images conditioned on pseudo labels, which are obtained from a clustering head $\mathcal{C}(\cdot)$ over the features from $\mathcal{F}(\cdot)$. 

To secure the model training, we deploy two-stage optimization mechanism a) \textbf{Stage I}: pre-training the clustering cornerstone; b) \textbf{Stage II}: fine-tuning with our newly-developed GCC.

% This section first introduces two basic modules, i.e., clustering network and image generation network as well as their pre-training details. After training them, the sequential part describes how to utilize their collaboration with category-wise matching to learn discriminative representation. 

\subsection{Stage I: Clustering Cornerstone}
\subsubsection{Contrastive Representation Learning} 
\label{section:cls}

Intuitively, semantically similar samples are expected to have very closer or even the same representation after $\mathcal{F}(\cdot)$. To achieve such a vision, the superior initialization of parameters $\mathcal{F}(\cdot)$ significantly facilitates to capture valuable object knowledge. Hence, self-supervised learning \cite{chen2020simple} became the primary initialization fashion on solving deep clustering problem, especially for the successful exploration of contrastive learning framework. This one indicates that the important semantics should be invariant to various transformations over the original visual signals and builds on this point to provide each data point with the corresponding unsupervised representation. Specifically, for arbitrary one image, contrastive learning solutions typically construct a positive pair by rotating or cropping it and randomly collect others from dataset to be its negative partners. To this end, minimizing the ratio of the distance of positive pairs over that of negative ones naturally learns transformation-invariant representation and reasonably initializes network parameter. Mathematically, this procedure is achieved with the following constraint over the high-level features:
\begin{equation}
    \label{cl_loss}
    \small
    \mathcal{L}_{clr}=-\sum_{i=1}^{B}\log \frac{\exp\large(\mathbf{sim}(\mathcal{F}(\mathbf{X}_{i}), \mathcal{F}(\mathbf{X}_{i}^{*}))/ \tau \large)}{\sum_{j=1, j\neq i}^{2B}\exp \large(\mathbf{sim}(\mathcal{F}(\mathbf{X}_{i}), \mathcal{F}(\mathbf{X}_{j}))/ \tau \large)},
\end{equation}
where $\mathbf{X}_{i}^{*}$ means adding transformations on $\mathbf{X}_{i}$, $\mathbf{sim}(a,b)$ measures the similarity between $a$ and $b$ such as cosine distance, $\tau$ is a adjustable parameter and $B$ is the batch size. Eq. (\ref{cl_loss}) explicitly suggests that there is one positive pair and $2B-1$ negative pairs for each selected sample. Considerable negative pairs are helpful to attain the tighter lower bound of mutual information and generate classification boundary among different clusters. 

\subsubsection{Pseudo-Label Assisted Clustering}

To initialize the clustering head $\mathcal{C}(\cdot)$, we first consider directly using K-means \cite{macqueen1965some} to divide these learned features into distinct categories. After the assignment, each sample $\mathbf{X}_{i}$ is annotated with the corresponding pseudo one-hot label $\mathbf{y}_{i}$. In this case, as Figure \ref{framework} shows, an additional clustering head $\mathcal{C}(\cdot)$ is connected with the pre-trained feature extractor. To inherit prior knowledge, the random initialized clustering head is optimized with the assistance of pseudo cluster labels:
\begin{equation}
    \label{ce_loss}
    \min_{\mathcal{C}} \sum\nolimits_{i} \mathcal{L}_{ce}(\mathbf{p}_{i}, \mathbf{y}_{i}),
\end{equation}
where $\mathcal{L}_{ce}$ is the cross-entropy loss with \textit{Softmax} activation on the logits $\mathbf{p}_{i}=\mathcal{C}(\mathcal{F}(\mathbf{X}_{i}))\in \mathbb{R}^{K}$ with $K$ clusters. It is worth noting that $\mathcal{F}(\cdot)$ is frozen while updating $\mathcal{C}(\cdot)$. 

So far, the clustering branch with feature extractor and clustering head has obtained the meaningful initialization via Eq. (\ref{cl_loss}) and Eq. (\ref{ce_loss}). However, the current feature representations obviously are not the optimal ones due to the inevitable defect of contrastive learning. The drawback mainly results from the random selection of negative partners. For example, in fact, the half of negative pairs are supposed to be positive ones when there only exist two categories. The optimization of Eq. (\ref{cl_loss}) likely expands the intra-class distance to reduce the discriminative ability of features. This problem also occurs in imbalanced category scenario. To break down the limitation, the intuitive consideration is to generate the demand-specific samples to achieve the downstream task.

\subsubsection{Conditional Diffusion Generation}

In fact, the above demand of sample generation needs a controllable procedure. In other words, given the concrete annotation, the generative model learns the intrinsic distribution of image belonging to this category and utilizes these properties to derive novel samples with high fidelity and diversity. Fortunately, the appealing denoising diffusion probabilistic model (DDPM) \cite{ho2020denoising} satisfies such requirements. DDPM framework includes the forward process and reverse one. The former one aims to diffuse the original images into a Gaussian sphere by gradually adding noise into visual signals, which is formulated as a Markov chain:
\begin{equation}
    \label{ddpm_forward}
q(\mathbf{X}_{i,1:T}|\mathbf{X}_{i,0})=\Pi_{t=1}^{T}q(\mathbf{X}_{i,t}|\mathbf{X}_{i,t-1}),
\end{equation}
where $q(\mathbf{X}_{i,t}|\mathbf{X}_{i,t-1})=\mathcal{N}\Large(\sqrt{1-\beta_{t}}\mathbf{X}_{i,t}, \beta_{t}\mathbf{I})$ with $\beta_{t}$ as a positive constant and $\mathbf{X}_{i,t}$ as the $t$-step signal of $\mathbf{X}_{i}$. With the chain conduction theory, $\mathbf{X}_{i,t}$ follows the distribution $q(\mathbf{X}_{i,t}|\mathbf{X}_{i,0})=\mathcal{N}\Large(\mathbf{X}_{i,t}; \sqrt{\alpha_{t}}\mathbf{X}_{i,0}, (1-\alpha_{t})\mathbf{I})$ where $\alpha_{t}=\Pi_{j=1}^{t}(1-\beta_{j})$. Hence, it is simple to deduce $\mathbf{X}_{i,t}=\sqrt{\alpha_{t}}\mathbf{X}_{i,0}+(1-\alpha_{t})\epsilon$ with $\epsilon\sim \mathcal{N}(\mathbf{0}, \mathbf{I})$. 

Inversely, the latter process attempts to estimate the added noise per step and remove it from $\mathbf{X}_{i,t}$ to recover $\mathbf{X}_{i,t-1}$ and finally obtains $\mathbf{X}_{i,0}$. Considering the conditional generation, we further advance DDPM by integrating discriminative feature matching into the denoising process. Specifically, given the frozen clustering branch, each real image $\mathbf{X}_{i}$ is first transformed to the cluster assignment with one-hot encoding $\hat{\mathbf{y}}_{i}\leftarrow \mathbf{p}_{ij}$. With $\hat{\mathbf{y}}_{i}$, one step of the denoising stage can be defined as:
\begin{equation}
    \label{ddpm_reverse}
    \small
    p_{\mathcal{G}}(\mathbf{X}_{i,t-1}|\mathbf{X}_{i,t}, \hat{\mathbf{y}}_{i})=\mathcal{N}(\mathbf{X}_{i,t-1}; \mu_{\mathcal{G}}(\mathbf{X}_{i,t}, \hat{\mathbf{y}}_{i}, t), \sigma(t)\mathbf{I}).
\end{equation}
Similar to DDPM, the specific parameterization of Eq. (\ref{ddpm_reverse}):
\begin{equation}
    \label{mu_sigma}
    \small
    \mu_{\mathcal{G}}(\mathbf{X}_{i,t}, \hat{\mathbf{y}}_{i}, t)=\frac{1}{\alpha_{t}}( \mathbf{X}_{i,t}-\frac{\beta_{t}}{\sqrt{1-\alpha_{t}}}\epsilon_{\mathcal{G}}(\mathbf{X}_{i,t}, \hat{\mathbf{y}}_{i}, t)),
\end{equation}
where $\epsilon_{\mathcal{G}}$ is the function of estimating noise with inputs $\{\mathbf{X}_{i,t}, \hat{\mathbf{y}}_{i}, t\}$. Note that $\mathcal{G}(\cdot)$ in Eq. (\ref{ddpm_reverse}) and (\ref{mu_sigma}) denotes the trainable network parameters. Moreover, we also have $\sigma(t)=(\frac{1-\alpha_{t-1}}{1-\alpha_{t}}\beta_{t})^{\frac{1}{2}}$. Finally, the key point is to achieve agreement between added noise and the estimated one via:
\begin{equation}
    \label{ddpm_loss}
    \min_{\mathcal{G}} \mathcal{L}_{g}=\mathbb{E}_{\mathbf{X}_{i}\sim q(\mathbf{X}_{i}), \epsilon\sim \mathcal{N}(\mathbf{0}, \mathbf{I})}\|\epsilon-\epsilon_{\mathcal{G}}(\mathbf{X}_{i,t}, \hat{\mathbf{y}}_{i}, t)\|_{2}^{2},
\end{equation}
which indicates that the annotation information affects the estimation of noise so that each category tends to have the unique accumulated noise pattern. After training the diffusion network $\epsilon_{\mathcal{G}}(\cdot)$, this property effectively guarantees that the model deciphers our interested semantics from random noise given the specific cluster label. Certainly, the quality of generation is positively associated with clustering accuracy. Thus, the remaining challenge is better utilizing these generated images further to promote the performance of clustering network and form collaboration between them to amplify the benefit.

\subsection{Stage II: Generative Calibration Clustering}

\subsubsection{Discriminative Feature Matching}

As \cite{rombach2022high} claims, the conventional DDPM actually observes and summarizes the statistical pattern of data population. Similarly, due to the guidance of annotation, the novel images from our conditional generation model basically reflect the core attribution of each category. This analysis suggests that the cluster centers computed from novel samples with annotations tend to be the ideal approximation of the realistic ones. Thus, we consider these centroids of novel instances as the benchmarks to adjust that of original images and formulate this distance in the reproducing kernel Hilbert space (RKHS) \cite{chen2016error} as:
\begin{equation}
    \label{centers}
    \forall k, \quad \sup_{\mathcal{F}\sim\mathcal{H}}(\mathbb{E}_{\mathbf{X}\sim q_{k}}[\mathcal{F}(\mathbf{X})]-\mathbb{E}_{\widetilde{\mathbf{X}}\sim \widetilde{q}_{k}}[\mathcal{F}(\widetilde{\mathbf{X}})])_{\mathcal{H}},
\end{equation}
where $q_{k}$ and $\widetilde{q}_{k}$ are the distributions of the $k$-cluster in original and novel datasets, and $\mathcal{H}$ denotes the RKHS.

Before instantiating Eq. (\ref{centers}), we first describe how to operate two branches to obtain the desired features and which components will be optimized or frozen. On the one hand, a batch of real images $\{{\mathbf{X}_{i}\}}_{i=1}^{B}$ is randomly chosen from the original dataset and fed into the clustering network to obtain the corresponding ${\{\mathcal{F}(\mathbf{X}_{i}}),\mathbf{p}_{i}, \hat{\mathbf{y}}_{i}\}_{i=1}^{B}$.
% ${\{\mathcal{F}(\mathbf{X}_{i}}), \eta(\mathbf{p}_{i}), \hat{\mathbf{y}}_{i}\}_{i=1}^{B}$ where $\eta(\cdot)$ is the \textit{Softmax} function.
On the other hand, a batch of random noise $\{\widetilde{\mathbf{X}}_{j,T}\in \mathbb{R}^{c\times h \times h}\}_{j=1}^{B}$ is sampled from normal distribution and the frozen denoising model takes them and the annotations $\hat{\mathbf{y}}_{j}$ as input to recover the fake images $\{\widetilde{\mathbf{X}}_{j,0}\}_{j=1}^{B}$ via Eq. (\ref{mu_sigma}). Subsequently, the clustering model converts these generative images into feature and cluster space as ${\{\mathcal{F}(\widetilde{\mathbf{X}}_{j}}),\widetilde{\mathbf{p}}_{j}, \widetilde{\mathbf{y}}_{j}\}_{j=1}^{B}$. Note that $\hat{\mathbf{y}}_{j}$ can be regarded as the ground-truth label for this generative sample. With these contents, it becomes convenient to achieve the instantiation. Concretely, for the $k$-th cluster, Eq. (\ref{centers}) can be rewritten as follows:
\begin{equation}
    \label{centers_v1}
    \|\sum_{i=1}^{B}\mathcal{I}_{(\hat{\mathbf{y}}_{i}\rightarrow{k})}\mathcal{F}(\mathbf{X}_{i})-\sum_{j=1}^{B}\mathcal{I}_{(\hat{\mathbf{y}}_{j}\rightarrow{k})}\mathcal{F}(\widetilde{\mathbf{X}}_{j,0})\|_{\mathcal{H}}^{2},
\end{equation}
where $\mathcal{I}_{(\hat{\mathbf{y}}_{j}\rightarrow{k})}$ is the sign function being 1 when the label $\mathbf{y}_{j}$ is $k$ ($\hat{\mathbf{y}}_{j}\rightarrow{k}$), otherwise, 0. Due to the computational efficiency of kernel function, Eq. (\ref{centers_v1}) is upgraded as:
\begin{eqnarray}
\label{centers_v2}
&\min\limits_{\mathcal{F}} \mathcal{L}_{d}=\sum\limits_{k=1}^{K} (d_{r}^{k}+d_{g}^{k}-2d_{rg}^{k}),~\mathrm{w.r.t}, \nonumber \vspace{4mm} \\
&\left\{
\begin{array}{lr}
d_{r}^{k} =\sum_{i=1}^{B}\sum_{j=1}^{B}\frac{\mathcal{I}_{(\hat{\mathbf{y}}_{i}\rightarrow{k})}\mathcal{I}_{(\hat{\mathbf{y}}_{j}\rightarrow{k})}\mathcal{K}(\mathcal{F}(\mathbf{X}_{i}), \mathcal{F}(\mathbf{X}_{j}))}{\sum_{i=1}^{B}\sum_{j=1}^{B}\mathcal{I}_{(\hat{\mathbf{y}}_{i}\rightarrow{k})}\mathcal{I}_{(\hat{\mathbf{y}}_{j}\rightarrow{k})}}, \vspace{2mm}\\
d_{g}^{k} =\sum_{i=1}^{B}\sum_{j=1}^{B}\frac{\mathcal{I}_{(\hat{\mathbf{y}}_{i}\rightarrow{k})}\mathcal{I}_{(\hat{\mathbf{y}}_{j}\rightarrow{k})}\mathcal{K}(\mathcal{F}(\widetilde{\mathbf{X}}_{i}), \mathcal{F}(\widetilde{\mathbf{X}}_{j}))}{\sum_{i=1}^{B}\sum_{j=1}^{B}\mathcal{I}_{(\hat{\mathbf{y}}_{i}\rightarrow{k})}\mathcal{I}_{(\hat{\mathbf{y}}_{j}\rightarrow{k})}}, \vspace{2mm}\\
d_{rg}^{k} =\sum_{i=1}^{B}\sum_{j=1}^{B}\frac{\mathcal{I}_{(\hat{\mathbf{y}}_{i}\rightarrow{k})}\mathcal{I}_{(\hat{\mathbf{y}}_{j}\rightarrow{k})}\mathcal{K}(\mathcal{F}(\mathbf{X}_{i}), \mathcal{F}(\widetilde{\mathbf{X}}_{j}))}{\sum_{i=1}^{B}\sum_{j=1}^{B}\mathcal{I}_{(\hat{\mathbf{y}}_{i}\rightarrow{k})}\mathcal{I}_{(\hat{\mathbf{y}}_{j}\rightarrow{k})}},
\end{array}
\right.
\end{eqnarray}
% where $d_{r}^{k} =\sum_{i=1}^{B}\sum_{j=1}^{B}\frac{\mathcal{I}_{(\hat{\mathbf{y}}_{i}=k)}\mathcal{I}_{(\hat{\mathbf{y}}_{j}=k)}\mathcal{K}(\mathcal{F}(\mathbf{X}_{i}), \mathcal{F}(\mathbf{X}_{j}))}{\sum_{i=1}^{B}\sum_{j=1}^{B}\mathcal{I}_{(\hat{\mathbf{y}}_{i}=k)}\mathcal{I}_{(\hat{\mathbf{y}}_{j}=k)}}, 
% d_{g}^{k} =\sum_{i=1}^{B}\sum_{j=1}^{B}\frac{\mathcal{I}_{(\hat{\mathbf{y}}_{i}=k)}\mathcal{I}_{(\hat{\mathbf{y}}_{j}=k)}\mathcal{K}(\mathcal{F}(\widetilde{\mathbf{X}}_{i}), \mathcal{F}(\widetilde{\mathbf{X}}_{j}))}{\sum_{i=1}^{B}\sum_{j=1}^{B}\mathcal{I}_{(\hat{\mathbf{y}}_{i}=k)}\mathcal{I}_{(\hat{\mathbf{y}}_{j}=k)}}, 
% d_{rg}^{k} =\sum_{i=1}^{B}\sum_{j=1}^{B}\frac{\mathcal{I}_{(\hat{\mathbf{y}}_{i}=k)}\mathcal{I}_{(\hat{\mathbf{y}}_{j}=k)}\mathcal{K}(\mathcal{F}(\mathbf{X}_{i}), \mathcal{F}(\widetilde{\mathbf{X}}_{j}))}{\sum_{i=1}^{B}\sum_{j=1}^{B}\mathcal{I}_{(\hat{\mathbf{y}}_{i}=k)}\mathcal{I}_{(\hat{\mathbf{y}}_{j}=k)}}$, 
where $\mathcal{K}(\cdot, \cdot)$ is the kernel function. This constraint effectively calibrates the cluster centers of real data along the correct direction by optimizing the feature extractor.

From another viewpoint, when associating real and generative samples via Eq. (\ref{centers_v2}), they have actually built a mirror image relationship. And it is straightforward to obtain the annotations of novel samples. Under this situation, adjusting the intra-class compactness and inter-class discrimination over the generative images positively affects the boundary of clusters to assisting model in capturing more discriminative features from real images. This manner also overcomes the main shortcoming of contrastive learning. Thus, we further modify Eq. (\ref{cl_loss}) as the novel discriminative feature matching:
\begin{eqnarray}
    \label{cwm}
    \min_{\mathcal{F}} \mathcal{L}_{cwm}=&\frac{\sum_{i=1}^{B}\sum_{j=1}^{B}\mathcal{I}_{(\hat{\mathbf{y}}_{i}\neq\hat{\mathbf{y}}_{j})} \mathbf{sim}(\mathcal{F}(\widetilde{\mathbf{X}}_{i}), \mathcal{F}(\widetilde{\mathbf{X}}_{j}))}{\sum_{i=1}^{B}\sum_{j=1}^{B}\mathcal{I}_{(\hat{\mathbf{y}}_{i}\neq\hat{\mathbf{y}}_{j})}} \nonumber \vspace{2mm}\\
    &-\frac{\sum_{i=1}^{B}\sum_{j=1}^{B}\mathcal{I}_{(\hat{\mathbf{y}}_{i}=\hat{\mathbf{y}}_{j})} \mathbf{sim}(\mathcal{F}(\widetilde{\mathbf{X}}_{i}), \mathcal{F}(\widetilde{\mathbf{X}}_{j}))}{\sum_{i=1}^{B}\sum_{j=1}^{B}\mathcal{I}_{(\hat{\mathbf{y}}_{i}=\hat{\mathbf{y}}_{j})}},
\end{eqnarray}
where the first term reduces the similarity among samples from different categories to expand inter-class distance, while the second one narrows the representation difference of similar images to form tighter intra-class space.

\subsubsection{Reliable Self-Supervised Clustering}

Apart from category-wise matching, using the annotations of generative sample as the strong supervisions to optimize clustering branch is certainly an intuitive manner to improve discriminative ability of features. However, it is difficult to guarantee the objects of generative images $\widetilde{\mathbf{X}}_{i}$ to be always consistent with the given semantics of category $\hat{\mathbf{y}}_{i}$, especially for the initial generation stage. Under this condition, the strict supervised signals is likely to mislead and confuse the feature learning. To avoid this problem, we propose a novel reliable metric learning loss as the following:
\begin{equation}
    \label{metric}
    \min_{\mathcal{F},~\mathcal{C}} \mathcal{L}_{ml}=\sum_{i=1}^{B}\sum_{k=1}^{K}\mathcal{I}_{(\widetilde{\mathbf{p}}_{ik}>\widetilde{\mathbf{p}}_{ij})}(\widetilde{\mathbf{p}}_{ik}-\widetilde{\mathbf{p}}_{ij})^{2}, 
    % j=\arg\max \hat{\mathbf{y}}_{i},
\end{equation}
where $\widetilde{\mathbf{p}}_{ik}$ means the $k$-th element of $\widetilde{\mathbf{p}}_{i}=\mathcal{C}(\mathcal{F}(\widetilde{\mathbf{X}}_{i}))$, and $j=\arg\max_{j}\hat{\mathbf{y}}_{ij}$ suggests that this novel sample comes from the $j$-th cluster. Eq. (\ref{metric}) not only facilitates model to yield discriminative predictions but also allows multiple possibilities. First, when $\mathcal{L}_{ml}$ achieves the minimum, it indicates that $\widetilde{\mathbf{p}}_{ij}\geq \widetilde{\mathbf{p}}_{ik}$ and the semantics of the $j$-th category occupy the dominant position in feature $\mathcal{F}(\widetilde{\mathbf{X}}_{i})$. Second, if $\widetilde{\mathbf{p}}_{ik}\geq \widetilde{\mathbf{p}}_{ij}$, our proposed metric preserves the semantics of the $k$-th cluster by only reducing it to $\widetilde{\mathbf{p}}_{ij}$.

\subsection{Model Training Flow}

As we mentioned, our deep clustering framework involves two stages: pre-training and calibrated clustering. During pre-training stage, we first adopt Eq. (\ref{cl_loss}) to train feature extractor and then froze it to optimize the clustering head via Eq. (\ref{ce_loss}). For high-quality generation, while fixing the pre-trained clustering branch, the generative diffusion network is updated with Eq. (\ref{ddpm_loss}). In the next stage, with the frozen generative model, our generative calibration clustering (GCC) further adjusts clustering network via:
\begin{equation}
    \label{all_loss}
    \min_{\mathcal{F},~\mathcal{C}} \mathcal{L}_{d}+\mathcal{L}_{cwm}+\mathcal{L}_{ml}.
\end{equation}

Benefited from the improvement of clustering accuracy, the real images and pseudo labels are utilized to fine-tune the generative network to promote the quality of fake samples via Eq. (\ref{ddpm_loss}). The collaborative updating of two branches will be repeated for many rounds.

\begin{table*}[t]
\centering
\caption{Comparisons of clustering performance of multiple methods on three challenging object image benchmarks. The best result is highlighted with \textbf{bold} type, while the second best is marked with \underline{underline}.}
% \ZD{Highlight?}}
\label{table:clustering_results}
\setlength{\tabcolsep}{10pt} % Default value: 6pt 
\renewcommand{\arraystretch}{1.1} % Default value: 1 
\begin{tabular}{c|ccc|ccc|cccc}
  \Xhline{1pt}
Datasets &\multicolumn{3}{c}{Cifar-10} &\multicolumn{3}{|c}{Cifar-100}	
&\multicolumn{3}{|c}{STL-10}\\ \hline
Metrics	&NMI &ACC &ARI &NMI &ACC &ARI &NMI &ACC &ARI\\   \Xhline{1pt}
k-means	\cite{macqueen1965some} &0.087	&0.229	&0.049	&0.084	&0.13	&0.028	&0.125	&0.192	&0.061\\ \hline
SC	\cite{zelnik2004self} &0.103	&0.247	&0.085	&0.09	&0.136	&0.022	&0.098	&0.159	&0.048\\ \hline
AC	\cite{gowda1978agglomerative} &0.105	&0.228	&0.065	&0.098	&0.138	&0.034	&0.239	&0.332	&0.14\\ \hline
NMF	\cite{cai2009locality} &0.081	&0.19	&0.034	&0.079	&0.118	&0.026	&0.096	&0.18	&0.046\\   \Xhline{1pt}
AE	\cite{bengio2006greedy} &0.239	&0.314	&0.169	&0.1	&0.165	&0.048	&0.25	&0.303	&0.161\\ \hline
DAE	\cite{vincent2010stacked} &0.251	&0.297	&0.163	&0.111	&0.151	&0.046	&0.224	&0.302	&0.152\\ \hline
DCGAN	\cite{radford2015unsupervised} &0.265	&0.315	&0.176	&0.12	&0.151	&0.045	&0.21	&0.298	&0.139\\ \hline
DeCNN	\cite{zeiler2010deconvolutional} &0.24	&0.282	&0.174	&0.092	&0.133	&0.038	&0.227	&0.299	&0.162\\ \hline
VAE	\cite{kingma2013auto} &0.245	&0.291	&0.167	&0.108	&0.152	&0.04	&0.2	&0.282	&0.146\\ \hline
JULE	\cite{yang2016joint} &0.192	&0.272	&0.138	&0.103	&0.137	&0.033	&0.182	&0.277	&0.164\\ \hline
DEC	\cite{xie2016unsupervised} &0.257	&0.301	&0.161	&0.136	&0.185	&0.05	&0.276	&0.359	&0.186\\ \hline
DAC	\cite{chang2017deep} &0.396	&0.522	&0.306	&0.185	&0.238	&0.088	&0.366	&0.47	&0.257\\ \hline
ADC \cite{haeusser2019associative} &- &0.325 &- &- &0.160 &- &- &0.530 &-\\ \hline
DDC \cite{chang2019deep} &0.424 &0.524 &0.329 &- &- &- &0.371 &0.489 &0.267 \\ \hline
DCCM	\cite{wu2019deep} &0.496	&0.623	&0.408	&0.285	&0.327	&0.173	&0.376	&0.482	&0.262\\ \hline
IIC \cite{ji2019invariant} &- &0.617 &- &- &0.257 &- &- &0.610 &- \\ \hline
PICA	\cite{huang2020deep} &0.591	&0.696	&0.512	&0.31	&0.337	&0.171	&0.611	&0.713	&0.531\\ \hline
DCDC \cite{dang2021doubly} &0.585 &0.699 &0.506 &0.310 &0.349 &0.179 &0.621 &0.734 &0.547\\   \Xhline{1pt}
Pretext+Kmeans \cite{chen2020simple} &0.598	&0.659	&0.509	&0.402	&0.395	&0.239	&0.604	&0.658	&0.506\\ \hline
SCAN	\cite{van2020scan} &\textbf{0.796}	&\underline{0.861}	&\underline{0.75}	&\underline{0.485}	&\underline{0.483}	&0.314	&\underline{0.703}	&\underline{0.818}	&\underline{0.661}\\ \hline
NNM	\cite{dang2021doubly} &0.748	&0.843	&0.709	&0.484	&0.477	&\underline{0.316}	&0.694	&0.808	&0.650\\ \hline
EDESC	\cite{cai2022efficient} &0.464	&0.627	&-	&0.37	&0.385	&-	&0.687	&0.745	&- \\ \hline
LNSCC	\cite{ma2022locally} &0.713	&0.820	&-	&0.446	&0.439	&-	&0.663	&0.738	&-\\ \hline
Ours &\underline{0.792}	&\textbf{0.904}	&\textbf{0.799}	&\textbf{0.527}	&\textbf{0.532}	&\textbf{0.358}	&\textbf{0.759}	&\textbf{0.866}	&\textbf{0.723}\\   \Xhline{1pt}
\end{tabular}\vspace{-3mm}
\end{table*}

\section{Experiments}

\subsection{Experimental Setup}

\noindent $\Diamond$ \textbf{Datasets.} \textbf{Cifar-10} \cite{krizhevsky2009learning} is a natural image dataset including 60,000 colour images (32$\times$32$\times$3) with 10 classes such as airplane, bird and dog, etc. Each category consists of 5,000 training images and 1,000 ones for testing. In fact, another clustering benchmark \textbf{Cifar-100} and Cifar-10 are the same dataset. Differently, these images are further divided into 100 fine-grained categories. In addition, \textbf{STL-10} is a smaller dataset from ImageNet \cite{deng2009imagenet} and involves 10 categories where each one includes 500 training samples and 800 test ones. Besides, it has 100,000 images belonging to multiple unknown classes. For the utilization of these three datasets, we follow the protocol of \cite{van2020scan,dang2021doubly} which trains model with the training data and evaluates the performance on test set. Moreover, for Cifar-100, 20 super-classes are considered as the ground-truth in the practical experiments.

\noindent $\Diamond$ \textbf{Baselines \& Metric.} The competed baselines mainly involve three categories. The first one is the conventional clustering strategy with low-level features, such as K-means \cite{macqueen1965some}, SC \cite{zelnik2004self}, AC \cite{gowda1978agglomerative}, NMF \cite{cai2009locality}. Differently, the second one is almost based on deep learning framework as AE \cite{bengio2006greedy}, DAE \cite{vincent2010stacked}, DCGAN \cite{radford2015unsupervised}, DeCNN \cite{zeiler2010deconvolutional}, VAE \cite{kingma2013auto}, JULE \cite{yang2016joint}, DEC \cite{xie2016unsupervised}, DAC \cite{chang2017deep}, ADC \cite{haeusser2019associative}, DCCM \cite{wu2019deep}, PICA \cite{huang2020deep} and EDESC \cite{cai2022efficient}. The third one is built on the pre-trained network architecture as Pre-text+K-means \cite{chen2020simple}, SCAN \cite{van2020scan}, NNM \cite{dang2021doubly} and LNSCC \cite{ma2022locally}. Our proposed method belongs to the third category.

For clustering task, there are three widely-used evaluation metrics, i.e., Accuracy (Acc), Normalized Mutual Information (NMI) and Adjusted Rand Index (ARI). The changing scale of three metrics is (0,1). The higher value per metric suggests the better clustering performance. Our method utilizes the output of clustering head as the clustering assignment.

\noindent $\Diamond$ \textbf{Implementation Details.} The clustering branch in our GCC adopts ResNet-18 as the feature extractor and two fully-connected layers as the clustering head, while the denoising diffusion generative network is a U-Net architecture \cite{ho2020denoising} which involves 16 residual convolutional block and each block has two convolutional layers with dropout and ReLU function. And the input of U-Net consists of visual signal and embedding of time and annotation. Clustering head and denoising networks both use Adam as optimizer with learning rates as 1e-4 and 2e-4, respectively, and are separately pre-trained about 100 and 300 epochs. After that, we adopt iterative manner to update two branches with 5 and 2 epochs in each round.

\subsection{Comparison Result}

Table \ref{table:clustering_results} summarizes and reports the clustering performance of multiple methods on three popular benchmarks. To make fair comparisons, the results of baselines are directly referred from the corresponding works. According to the comparisons, we easily achieve several conclusions.

\textbf{First}, compared with the existing clustering methods, our GCC outperforms them by a significant margin. For example, on Cifar-10, our method surpasses the second one (SCAN) by 4.3\% on accuracy. It illustrates the advantage of our proposed GCC on solving deep clustering problem.

\textbf{Second}, although Cifar-10 and Cifar-100 are the identical datasets, all methods suffer from considerable performance degradation when evaluated from the former to the latter. The main reason lies in that the clustering problem becomes more difficult as the increasing number of categories. For these challenges, our method still obtains better clustering accuracy than others, which further verifies the effectiveness of our method on learning discriminative representations for downstream tasks. 

\begin{figure*}[t]
    \centering
    \includegraphics[width=1.0\linewidth]{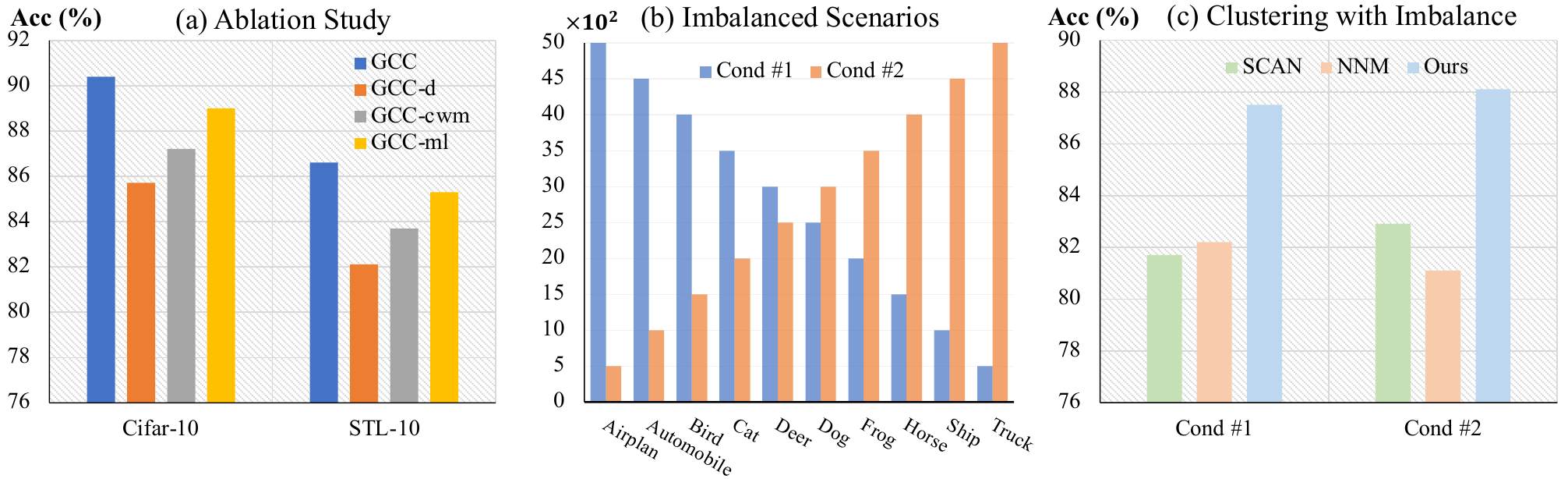}\vspace{-1mm}
    \caption{(a) Ablation study analyzes the effect of each loss function. (b) Statistics of imbalanced data distribution. (c) Clustering performance of three methods under imbalanced scenarios.}\vspace{-3mm}
    \label{ablation}
\end{figure*}

\textbf{Third}, when competing with traditional clustering methods with low-level features, the deep neural network based solutions show extensive superiority. The success of deep clustering mainly results from that the deeper network architecture captures abundant abstract semantics from the high-dimensional images to better facilitate downstream tasks. Hence, the core of clustering is still conducting better unsupervised representation learning. In fact, Pre-text+K-means is exploring contrastive learning manner to extract meaningful feature representation for each data point and then utilizing K-means over these learned features to perform the assignment of clusters. Hence, this method is similar with our pre-training procedure. The comparison with it fully reflects the positive influence of introducing conditional generated images. These novel instances with annotations provides the relative strong supervised signals to calibrate the learning of clustering network and leads it into the correct direction. On the other hand, SCAN and NNM actually are both constructed on the pre-training mechanism and further fine-tunes the deep clustering model to capture more object-relevant semantics. Specially, NNM uses the neighborhood concept to delicately redefine contrastive learning loss function, while SCAN further minimizes intra-class distance to form compact feature space. Different from them, our proposed method pays more attention to the constraints over the hidden features of generated images and associates them with the realistic images via the calibration of category centers. This learning manner also brings additional benefits to attain performance improvement. 

\textbf{Finally}, generative model has assisted clustering task, such as DCGAN, which adopts generative adversarial network (GAN) to approximate the real data distribution of each category and combines them with real images to optimize clustering network. However, GAN easily occurs mode collapse, resulting in the lack of sample diversity. And it ignores the exploration of association between real and generative images. In terms of these two aspects, our method conducts effective advances by exploring the conditional diffusion model and the adjustment of class centroids, which triggers in the larger performance promotion over DCGAN.

\subsection{Empirical Analysis}

This section will discuss the effect of each loss function on model performance and report the clustering results on imbalanced scenarios. In addition, the visualization of generative images and feature representations are exhibited to deeply analyze the working mechanism of our GCC.

\begin{figure*}
    \centering
    \includegraphics[width=1.0\linewidth]{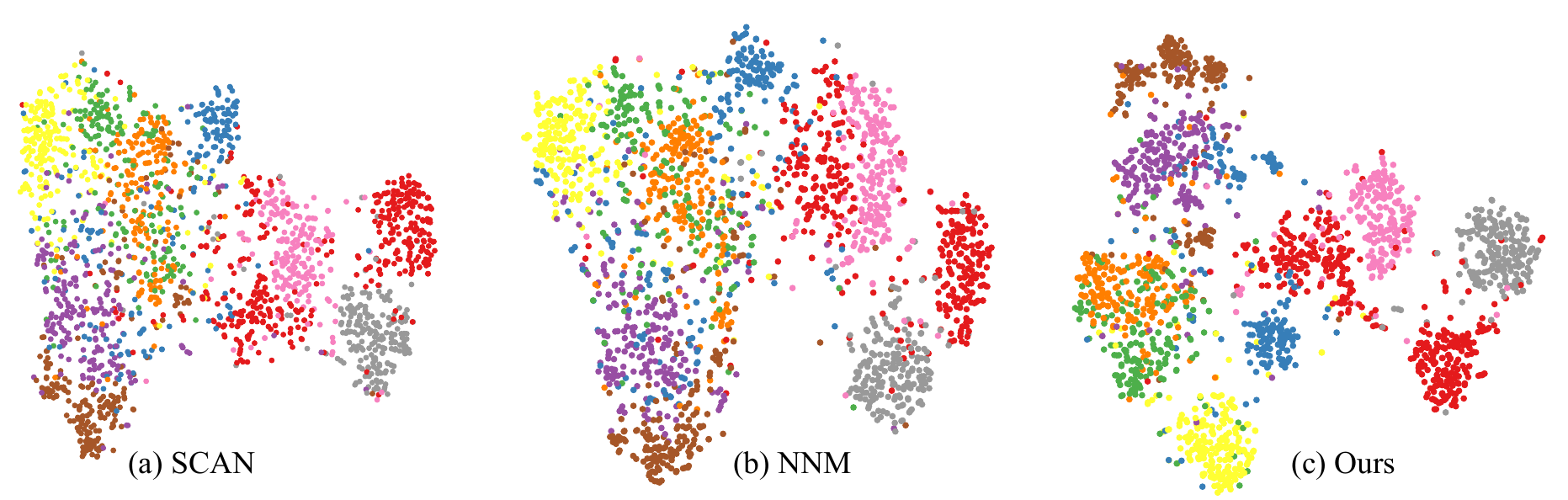}\vspace{-2mm}
    \caption{Visualization of hidden feature distribution from $\mathcal{F}(\cdot)$, where experiments are conducted on Cifar-10 dataset.}
    \label{feat_vis}\vspace{-3mm}
\end{figure*}

\begin{figure*}
    \centering
    \includegraphics[width=1\linewidth]{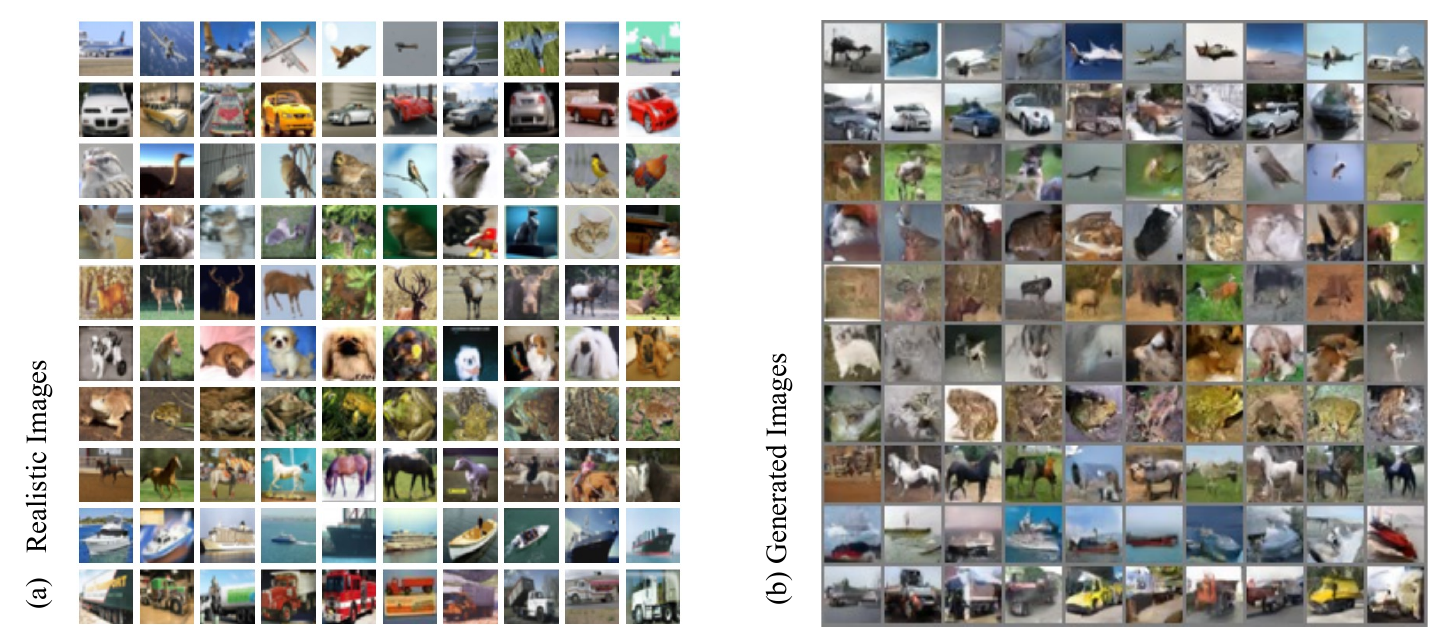}\vspace{-2mm}
    \caption{Comparison between realistic images and generative images generated by denoising network.}\vspace{-3mm}
    \label{images}
\end{figure*}

\subsubsection{Ablation \& Imbalanced Study}
% the effect of loss
There are three important loss functions $\mathcal{L}_{d}$, $\mathcal{L}_{cwm}$ and $\mathcal{L}_{ml}$ affecting the model learning. Concretely, $\mathcal{L}_{cwm}$ utilizes the annotation information of novel images derived by denoising model to accurately estimate the intra-class and inter-class similarity and refine feature space by narrowing or expanding them. Moreover, $\mathcal{L}_{ml}$ revises the logistics output $\widetilde{\mathbf{p}}_{i}$ of augmented instances to emphasize the class-relevant semantics and instruct model to produce more discriminative feature representations for each data point. Differently, the $\mathcal{L}_{d}$ focuses on constructing the association between generative and realistic images and transmits the effect of $\mathcal{L}_{cwm}$ and $\mathcal{L}_{ml}$ into feature learning of real instances. 

To clearly describe the contribution of each loss function, we create three variants of our GCC by separately removing $\mathcal{L}_{d}$, $\mathcal{L}_{cwm}$ and $\mathcal{L}_{ml}$ from Eq. (\ref{all_loss}) and naming them as GCC-d, GCC-cwm and GCC-ml. Theses variants are trained and evaluated on Cifar-10 and STL-10 with the results in Figure \ref{ablation} (a). It demonstrates that removing arbitrary one loss function leads to the reduction of clustering accuracy. In other words, these three constraints both have positive influence on learning better feature representations. Moreover, we find that without $\mathcal{L}_{d}$, the performance degradation is pretty significant. This observation means that only adding constraints over the generative images only supplies the limited assistance to cluster realistic samples. The main reason is that there are distribution shift across the novel augmented images and the real ones. Besides, the comparison between GCC-cwm and GCC-ml suggests that positive adjustment on feature distribution more effectively solve clustering issue.

On the other hand, when examining contrastive feature learning, Section \ref{section:cls} mentions and analyzes the defect of this learning manner on dealing with imbalanced data distribution. To observe the performance of our GCC under this scenario, we design two imbalanced clustering experiments on Cifar-10 dataset by randomly removing several samples per category with the statistics in Figure \ref{ablation} (b). The baselines (SCAN \& NNM) and our GCC are trained with these imbalanced datasets and evaluated on the original test set. From Figure \ref{ablation} (c), our method still achieves the higher clustering accuracy than others in the more challenging setting. And, compared with the results in Table \ref{table:clustering_results}, the performance of SCAN reduces about 5\%, while the drop of our GCC is only 3\%. The empirical studies also indicate that our GCC is relatively insensitive to imbalanced problem and better overcomes the limitation of contrastive learning.

\subsubsection{Feature Visualizations and Image Generation}
% Feature & Generative images.
For clustering task, the ideal situation is that the semantically similar images are embedded into the identical subspace, while instances with different object information lie in various clusters. Hence, besides quantitical metrics, comparing the feature distribution across multiple methods also reflects the performance of model. Given the well-trained model, we feed test images of Cifar-10 into the feature extractor to obtain their hidden representations from $\mathcal{F}(\mathbf{X_{i}})$. And then t-SNE \cite{van2008visualizing} is utilized to transform and visualize them in 2-D canvas. For the clear comparison, we randomly select 2,000 samples and show them in Figure \ref{feat_vis} where data points from distinct classes are painted with various colors. From the result, we find that the boundaries (learned by our GCC) among multiple categories are explicit and the intra-class distance is smaller than that of others. It further suggests that our GCC learns more discriminative features to solve clustering task.

Additionally, the merit of our method partly results from the introduce of novel augmented samples. High-quality generative images assist clustering model in obtaining more gains. Thus, we record several images generated by the conditional diffusion model during training stage. As each row of Figure \ref{images} (c) shows, the generative model indeed captures the basic property of category and recovers the desired images from random noise conditioned on the annotation. But there still exist obvious differences with the realistic images on color and resolution. Such differences lead to the feature distribution shift across these two sets. Thus, the constraint $\mathcal{L}_{d}$ plays a key role in improving feature learning and clustering performance.

\section{Conclusion}

In this paper, we present a novel Generative Calibration Clustering (GCC) method to address image clustering task. The basic motivation is to utilize the augmented image with annotation as powerful supervision to benefit feature learning of clustering. But empirical studies found that this strategy is not always valid due to the distribution shift across generative and real images. To overcome this issue, our GCC explores class centers of two sets to associate these two type of images. In addition, our GCC develops category-wise matching and self-supervised metric constraint to learn discriminative semantics. Experimental results and analysis both verify the advantage of our GCC.

\newpage
\balance
% {\small
% \bibliographystyle{ieee_fullname}
% \bibliography{egbib}
% }

\end{document}